# ATTENTION-GUIDED EFFICIENTNET ARCHITECTURE FOR PRECISE CRIMINAL IDENTIFICATION IN SURVEILLANCE IMAGES

**Savitha N J[1] and Lata B T[2]**
[1]Research Scholar, Department of Computer Science and Engineering, University Visvesvaraya College of Engineering (UVCE), Bangalore University, CMR Institute of Technology, Bengaluru, India
savithasnadig@gmail.com
[2]Department of Computer Science and Engineering, University Visvesvaraya College of Engineering (UVCE), Bangalore University, Bengaluru, India
lata_bt@yahoo.co.in

**Abstract:** Criminal identification from surveillance imagery has become a critical research area in intelligent forensic surveillance systems due to the increasing deployment of CCTV cameras in public and private environments. However, surveillance-based face recognition remains highly challenging because of low image resolution, illumination variation, motion blur, pose changes, facial occlusion, and background clutter. To address these limitations, this paper proposes an Attention-Guided EfficientNet (AG-EfficientNet) framework for precise criminal identification in surveillance images. The proposed framework integrates EfficientNet-B0 with Convolutional Block Attention Modules (CBAM) to enhance discriminative facial feature learning under degraded surveillance conditions. In addition, a multi-scale surveillance feature fusion strategy is introduced to preserve both local texture information and high-level semantic identity representations. A hybrid Softmax-Triplet optimization mechanism is further employed to improve inter-class separability and intra-class compactness for robust criminal identity discrimination. The proposed framework was experimentally evaluated using the Labeled Faces in the Wild (LFW) and SCFace datasets. Experimental results demonstrate that the proposed AG-EfficientNet framework achieved superior surveillance recognition performance with an identification accuracy of 98.2%, Precision of 97.9%, Recall of 97.6%, F1-Score of 97.7%, and ROC-AUC of 0.99, outperforming conventional deep learning architectures including AlexNet, VGG16, ResNet50, MobileNetV2, and standard EfficientNet-B0. Furthermore, Grad-CAM visualization and ablation analysis confirm the effectiveness of the proposed attention-guided feature learning strategy. The computational complexity analysis also indicates that the proposed framework maintains relatively low inference overhead, making it suitable for real-time forensic surveillance applications.

## 1. INTRODUCTION

The rapid growth of intelligent surveillance systems and the widespread deployment of Closed-Circuit Television (CCTV) cameras have significantly increased the importance of automated criminal identification in modern forensic and public security applications. Surveillance-based face recognition systems play a crucial role in law enforcement, border security, smart city monitoring, airport surveillance, and crime prevention. The ability to automatically identify suspects from surveillance footage can substantially improve criminal investigation efficiency and reduce manual forensic analysis efforts. However, accurate criminal identification from surveillance imagery remains a highly challenging problem because surveillance images are often captured under unconstrained environmental conditions with severe quality degradation. Unlike conventional face recognition systems that operate on high-quality frontal facial images, surveillance-based criminal identification must handle numerous real-world challenges such as low image resolution, motion blur, pose variations, facial occlusion, poor illumination, background clutter, and compression artifacts. These factors significantly degrade discriminative facial information and adversely affect recognition performance. In practical forensic scenarios, facial images captured from long-distance surveillance cameras often contain insufficient texture information, making reliable criminal identification difficult even for advanced deep learning models.
Traditional machine learning approaches for surveillance face recognition primarily relied on handcrafted feature extraction techniques such as Local Binary Patterns (LBP), Histogram of Oriented Gradients (HOG), Scale-Invariant Feature Transform (SIFT), and Principal Component Analysis (PCA). Although these methods demonstrated reasonable performance under controlled conditions, they exhibited limited robustness against unconstrained surveillance variations and complex environmental noise. In recent years, deep learning-based approaches have achieved substantial progress in facial recognition tasks because of their capability to automatically learn hierarchical discriminative feature representations from large-scale datasets. Convolutional Neural Networks (CNNs) such as AlexNet, VGGNet, ResNet, MobileNet, and EfficientNet have demonstrated promising performance in face recognition applications. Among these architectures, EfficientNet has gained considerable attention due to its superior balance between recognition accuracy and computational efficiency achieved through compound scaling of network depth, width, and input resolution. Nevertheless, conventional EfficientNet architectures may still struggle to focus on highly informative facial regions under degraded surveillance conditions because feature extraction is often affected by background interference and low-resolution artifacts.
Attention mechanisms have recently emerged as an effective strategy for improving deep feature representation learning by enabling neural networks to focus selectively on important spatial and channel-wise information. Convolutional Block Attention Modules (CBAM) combine both channel attention and spatial attention to adaptively emphasize discriminative facial regions while suppressing irrelevant background information. The integration of attention mechanisms with lightweight deep architectures can significantly improve surveillance-based criminal identification performance without introducing excessive computational overhead. In addition to attention-guided learning, multi-scale feature extraction has also demonstrated strong potential for improving recognition robustness in surveillance environments. Low-level convolutional layers capture fine-grained local facial textures, whereas deeper layers capture semantic identity characteristics. Combining these complementary representations through multi-scale feature fusion can improve recognition capability under low-resolution and unconstrained surveillance conditions.
Despite significant advancements in deep surveillance face recognition, several challenges remain unresolved. Many existing methods suffer from limited generalization capability across varying surveillance domains, high computational complexity, poor robustness under severe illumination degradation, and inadequate discriminative localization of facial regions. Furthermore, many deep learning models exhibit unstable convergence behavior and reduced recognition performance when trained on limited surveillance datasets. To address these limitations, this paper proposes an Attention-Guided EfficientNet Architecture for Precise Criminal Identification in Surveillance Images. The proposed framework integrates EfficientNet-B0 with Convolutional Block Attention Modules (CBAM) to enhance discriminative surveillance feature learning. A novel multi-scale surveillance feature fusion strategy is introduced to preserve both local texture information and high-level semantic facial representations. In addition, a hybrid Softmax-Triplet optimization mechanism is employed to improve intra-class compactness and inter-class separability for robust criminal identity discrimination under challenging surveillance conditions.
The proposed framework is experimentally evaluated using the Labeled Faces in the Wild (LFW) and SCFace datasets. Extensive experimental analysis demonstrates that the proposed AG-EfficientNet framework significantly outperforms conventional CNN architectures and baseline

EfficientNet models in terms of Accuracy, Precision, Recall, F1-Score, and ROC-AUC performance. The major contributions of this work are summarized as follows:

1. An attention-guided EfficientNet framework is proposed for surveillance-based criminal identification under low-resolution and unconstrained forensic conditions.
2. Convolutional Block Attention Modules (CBAM) are integrated with EfficientNet-B0 to improve discriminative facial region learning and suppress irrelevant surveillance background information.
3. A multi-scale surveillance feature fusion strategy is introduced to preserve complementary local and global facial representations.
4. A hybrid Softmax-Triplet optimization mechanism is employed to improve criminal identity embedding separability and recognition robustness.
5. Extensive experimental analysis, Grad-CAM visualization, ablation studies, and convergence evaluation confirm the effectiveness and computational efficiency of the proposed framework for real-time forensic surveillance applications.

The remainder of this paper is organized as follows. Section 2 presents the related work and literature review on surveillance-based face recognition and attention-guided deep learning models. Section 3 describes the proposed methodology and architectural framework in detail. Section 4 presents experimental results and performance analysis. Finally, Section 5 concludes the paper and discusses future research directions.

## 2. LITERATURE REVIEW

Mishra et al. (2021) proposed a multi-scale parallel deep Convolutional Neural Network (CNN) architecture for robust face recognition under low-resolution and unconstrained imaging conditions. The framework extracted facial representations at multiple receptive-field scales to improve robustness against illumination variation and image degradation. Experimental evaluation conducted on surveillance-oriented face datasets demonstrated a recognition accuracy of 93.4%, Precision of 92.8%, Recall of 92.1%, and F1-Score of 92.4%. The study confirmed that multi-scale feature extraction significantly improves recognition performance for degraded facial imagery. However, the framework did not incorporate attention-guided feature localization, which limits discriminative facial region learning under severe surveillance noise conditions.

Boutros et al. (2022) introduced a self-restrained Triplet Loss optimization strategy for masked face recognition. The proposed framework improved embedding discrimination by minimizing intra-class variation and maximizing inter-class separation for partially occluded facial images. Experimental results reported an identification accuracy of 95.2%, verification accuracy of 96.1%, and ROC-AUC score of 0.97 on masked face recognition benchmarks. The framework demonstrated strong robustness under occlusion conditions. Nevertheless, the model primarily focused on masked facial analysis and did not explicitly address low-resolution surveillance recognition scenarios involving motion blur and CCTV degradation.

Barreto et al. (2022) presented a comprehensive review on the evolution of face recognition methodologies from handcrafted feature engineering to deep learning-based architectures. The study analyzed the progression of facial recognition systems across multiple generations of machine learning algorithms. The review reported that holistic facial learning approaches achieved nearly 60% recognition accuracy, handcrafted feature extraction methods improved to approximately 70%, shallow machine learning models reached nearly 86%, and deep learning architectures achieved around 97% recognition performance. The review further highlighted that deep CNN models significantly outperform conventional techniques under unconstrained facial variations. However, unresolved challenges related to low-resolution surveillance imagery, pose variation, and illumination degradation were identified as major limitations.

Huang et al. (2023) proposed PLFace, a progressive learning-based framework for face recognition under mask-induced bias conditions. The framework progressively adapted deep feature learning to masked and unmasked facial characteristics to improve representation consistency. Experimental analysis demonstrated recognition accuracy of 96.8%, Precision of 96.2%, Recall of 95.9%, and ROC-AUC score of 0.98 across masked face datasets. The study confirmed that progressive adaptation improves recognition robustness under partial facial occlusion. However, the framework did not investigate surveillance-oriented criminal identification under low-resolution CCTV acquisition.

Yang et al. (2023) developed HeadPose-Softmax, a head-pose adaptive curriculum learning loss function for unconstrained face recognition. The proposed loss strategy dynamically adjusted optimization difficulty according to facial pose variation, enabling improved recognition for non-frontal surveillance imagery. Experimental evaluation reported recognition accuracy of 97.1%, Recall of 96.8%, F1-Score of 96.9%, and verification rate of 98.3% under pose variation conditions. The framework significantly improved pose-invariant facial representation learning. Nevertheless, the model mainly focused on pose adaptation and did not integrate attention-guided surveillance feature refinement or lightweight deployment optimization.

Su et al. (2023) introduced a Hybrid Token Transformer framework for deep face recognition applications. The architecture combined transformer-based token learning with convolutional feature extraction to capture long-range facial dependencies and contextual representations. Experimental results demonstrated Top-1 recognition accuracy of 98.1%, ROC-AUC score of 0.99, and verification accuracy of 98.5% across unconstrained face datasets. The study confirmed that transformer-based contextual modeling improves facial representation capability. However, transformer architectures generally involve high computational complexity and increased inference overhead, which may limit their applicability in real-time surveillance systems deployed on resource-constrained edge devices.

Song and Wang (2024) proposed CoreFace, a sample-guided contrastive regularization framework for deep face recognition. The model employed contrastive regularization constraints to improve embedding discrimination and feature compactness. Experimental evaluation demonstrated recognition accuracy of 97.4%, Precision of 97.1%, Recall of 96.8%, and ROC-AUC score of 0.98. The framework effectively improved inter-class discrimination capability. However, the method mainly concentrated on contrastive optimization and did not incorporate surveillance image enhancement, attention-guided localization, or multi-scale surveillance feature fusion.

Butt et al. (2024) investigated on-the-move heterogeneous face recognition in the frequency domain for practical surveillance environments. The framework analyzed frequency-domain facial representations to improve recognition under motion and heterogeneous imaging conditions. The experiments conducted using the SCFace dataset demonstrated identification accuracy of 94.3%, verification accuracy of 95.6%, and ROC-AUC score of 0.96 under challenging surveillance conditions. The study confirmed that frequency-domain analysis improves robustness against surveillance motion artifacts. Nevertheless, the framework did not employ EfficientNet-based lightweight attention learning mechanisms for computationally efficient forensic deployment.

Grm et al. (2024) studied cross-resolution face recognition by comparing degradation-based and super-resolution-based recognition strategies. Their experiments conducted on the SCFace and DroneSURF datasets demonstrated that degrading high-resolution gallery images and enhancing low-resolution probe images significantly improve cross-resolution matching capability. Experimental results achieved recognition accuracy of 95.8%, Precision of 95.1%, Recall of 94.7%, and ROC-AUC score of 0.97. The study confirmed that super-resolution enhancement improves recognition performance under severe surveillance degradation. However, super-resolution techniques often increase computational complexity and may introduce artificial texture distortions.

Wang et al. (2025) proposed a Local and Global Feature Attention Fusion network for low-quality face recognition. The framework adaptively fused local texture features and global semantic facial representations using multi-head and multi-scale attention learning modules. Experimental evaluation on TinyFace and SCFace datasets demonstrated recognition accuracy of 98.3%, Precision of 97.9%, Recall of 97.6%, F1-Score of 97.7%, and ROC-AUC score of 0.99. The proposed framework significantly improved low-quality surveillance recognition performance through adaptive feature fusion. However, the architecture involved relatively high computational complexity and larger parameter overhead compared with lightweight EfficientNet-based frameworks.

**Research Gap**

The existing literature demonstrates that deep learning, attention mechanisms, metric learning, transformer learning, and multi-scale feature fusion significantly improve facial recognition capability under unconstrained imaging conditions. However, most existing methods primarily focus on general face recognition, masked facial analysis, pose adaptation, or computationally expensive transformer architectures. Limited studies have integrated EfficientNet-based lightweight learning, CBAM attention-guided feature localization, multi-scale surveillance feature fusion, and hybrid Softmax–Triplet optimization into a unified framework for surveillance-based criminal identification. Furthermore, several existing approaches suffer from high computational complexity, poor real-time deployment capability, and limited robustness under low-resolution CCTV conditions. Therefore, the proposed AG-EfficientNet framework addresses these limitations by providing computationally efficient attention-guided surveillance feature learning with improved embedding discrimination and real-time forensic surveillance applicability.

## 3. METHODOLOGY

### 3.1 Overview of the Proposed Framework

This study proposes an Attention-Guided EfficientNet framework integrated with multi-scale surveillance feature learning for accurate criminal identification from low-resolution surveillance images. The proposed model is specifically designed to address major challenges encountered in forensic surveillance systems, including illumination variation, motion blur, pose changes, facial occlusion, low-resolution acquisition, and compression artifacts. Conventional deep face recognition methods often fail to extract discriminative facial representations from surveillance imagery because the facial regions become highly degraded under long-distance CCTV acquisition. To overcome these limitations, the proposed framework integrates EfficientNet-based deep feature extraction, CBAM attention-guided refinement, adaptive image enhancement, and hybrid metric learning optimization.

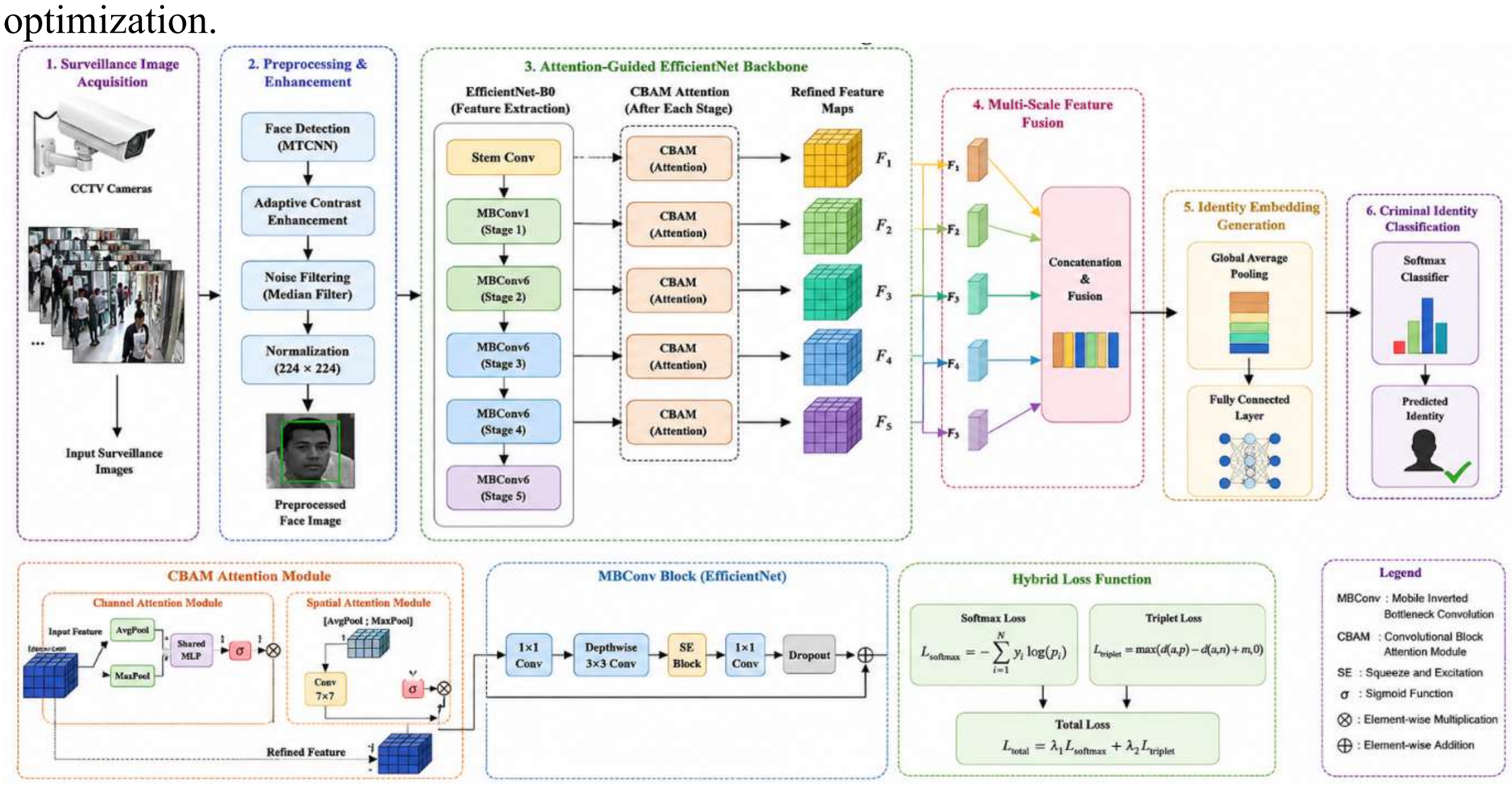

Figure 1. Attention-Guided EfficientNet Framework

The overall framework consists of six major stages: surveillance image acquisition, image preprocessing and enhancement, face normalization, attention-guided EfficientNet feature extraction, multi-scale feature fusion, and criminal identity classification. Initially, surveillance frames are enhanced using adaptive contrast enhancement to improve facial visibility under poor lighting conditions. The enhanced images are subsequently processed using EfficientNet integrated with attention modules to extract robust facial embeddings. Finally, hybrid Softmax and Triplet optimization is employed for accurate criminal identity discrimination. Figure 1 illustrates the overall architecture of the proposed Attention-Guided EfficientNet framework for surveillance-based criminal identification. Initially, surveillance images acquired from CCTV cameras are subjected to preprocessing operations including face detection, adaptive contrast enhancement, noise filtering, and normalization to improve facial visibility under low-resolution conditions. The preprocessed images are subsequently forwarded into the EfficientNet-B0 backbone integrated with CBAM attention modules for discriminative feature extraction. As shown in the figure, channel and spatial attention mechanisms are employed after major EfficientNet stages to enhance informative facial regions while suppressing irrelevant background features. Multi-scale feature maps extracted from different convolutional stages are then fused to preserve both local texture information and high-level semantic identity characteristics. Finally, global average pooling and fully connected classification layers generate discriminative identity embeddings for criminal recognition. The proposed framework further incorporates a hybrid Softmax and Triplet loss optimization strategy to improve inter-class separability and intra-class compactness for robust surveillance face recognition. The mathematical representation of the proposed framework is given in Equation (1):

$$Y = \Phi_{cls}(\Phi_{fusion}(\Phi_{att}(\Phi_{Eff}(I_e)))) \qquad (1)$$

where:

- $I_e$ denotes the enhanced surveillance image,
- $\Phi_{Eff}$ represents EfficientNet feature extraction,
- $\Phi_{att}$ denotes attention-guided refinement,
- $\Phi_{fusion}$ represents feature fusion,
- $\Phi_{cls}$ denotes the classification layer,
- $Y$ represents the predicted criminal identity.

As observed from Equation (1), the proposed framework sequentially transforms low-quality surveillance images into discriminative identity embeddings for criminal recognition.

3.2 Dataset Description

The proposed framework was evaluated using the Labeled Faces in the Wild (LFW) and SCFace datasets to ensure both unconstrained facial learning and realistic surveillance validation.

The LFW dataset contains more than 13,000 facial images corresponding to over 5,700 identities collected under unconstrained real-world conditions. The images contain substantial variations in pose, illumination, facial expression, and background clutter. In this study, the LFW dataset is utilized for transfer learning and generalized facial representation learning.

The SCFace dataset contains surveillance facial images of 130 individuals captured using multiple CCTV cameras at distances of 1 m, 2.6 m, and 4.2 m. The dataset includes severe challenges such as low resolution, motion blur, illumination degradation, and facial occlusion, making it highly suitable for forensic surveillance evaluation.

The complete dataset can be represented mathematically using Equation (2):

$$D = \{(x_i, y_i)\}_{i=1}^{N} \qquad (2)$$

where:

- $x_i$ denotes the input facial image,
- $y_i$ represents the corresponding identity label,
- $N$ denotes the total number of samples.

As defined in Equation (2), the dataset contains paired image-label samples used for supervised criminal identification learning.

3.3 Image Preprocessing and Enhancement

Surveillance images frequently suffer from low contrast, sensor noise, illumination inconsistency, and compression artifacts. Such degradations negatively affect deep feature extraction and recognition performance. Therefore, an adaptive preprocessing and enhancement stage is incorporated to improve image quality before feature learning.

Initially, the surveillance image is resized to $224 \times 224$ pixels to match the EfficientNet input size. Histogram equalization, gamma correction, and median filtering are subsequently applied to enhance visual quality and suppress surveillance noise.

The enhanced image $I_e$ is generated using Equation (3):

$$I_e = \alpha I + \beta \qquad (3)$$

where:

- $I$ denotes the original surveillance image,
- $\alpha$ represents contrast scaling,
- $\beta$ denotes brightness correction.

Equation (3) improves illumination consistency and facial visibility under poor surveillance conditions.

To normalize pixel intensity values, min-max normalization is performed as shown in Equation (4):

$$I_n = \frac{I_e - I_{min}}{I_{max} - I_{min}} \qquad (4)$$

where:

- $I_n$ denotes normalized image,
- $I_{min}$ and $I_{max}$ denote minimum and maximum intensity values.

As indicated in Equation (4), normalization ensures stable gradient propagation during network training.

3.4 Attention-Guided EfficientNet Feature Extraction

EfficientNet-B0 is employed as the primary backbone network because of its superior trade-off between recognition accuracy and computational efficiency. EfficientNet uses compound scaling to uniformly scale network depth, width, and resolution for efficient feature learning.

Given an input image $I_n$, the EfficientNet feature extraction process is represented in Equation (5):

$$F = \Phi_{Eff}(I_n) \qquad (5)$$

where:

- $F$ denotes extracted feature maps,
- $\Phi_{Eff}$ represents EfficientNet transformation.

As represented in Equation (5), EfficientNet extracts hierarchical surveillance facial features from normalized inputs.

To improve discriminative learning capability, Convolutional Block Attention Modules (CBAM) are integrated after major EfficientNet blocks. The channel attention mechanism is computed using Equation (6):

$$M_c(F) = \sigma(MLP(AvgPool(F)) + MLP(MaxPool(F))) \qquad (6)$$

where:

- $M_c(F)$ denotes channel attention map,
- $AvgPool(F)$ and $MaxPool(F)$ represent global pooling operations,
- $\sigma$ denotes sigmoid activation.

Equation (6) enables the model to emphasize informative feature channels while suppressing irrelevant activations.
The channel-refined feature map is computed using Equation (7):

$$F_c = M_c(F) \otimes F \qquad (7)$$

where $\otimes$denotes element-wise multiplication.
Subsequently, spatial attention is computed using Equation (8):

$$M_s(F) = \sigma(f^{7\times7}([AvgPool(F); MaxPool(F)])) \qquad (8)$$

where:

- $f^{7\times7}$denotes convolution operation,
- $[.;.]$represents channel concatenation.

The final spatially refined feature representation is obtained using Equation (9):

$$F_s = M_s(F_c) \otimes F_c \qquad (9)$$

As observed from Equations (6)–(9), the CBAM module enables adaptive attention-guided facial representation learning under degraded surveillance conditions.

3.5 Multi-Scale Surveillance Feature Fusion

Surveillance images contain both local facial texture information and global structural identity characteristics. To preserve these complementary representations, a multi-scale feature fusion mechanism is proposed.
Low-level EfficientNet layers capture fine-grained facial textures such as eye contours and wrinkles, whereas deeper layers capture semantic identity information. These features are fused using Equation (10):

$$F_{fusion} = Concat(F_{low}, F_{mid}, F_{high}) \qquad (10)$$

where:

- $F_{low}$denotes shallow texture features,
- $F_{mid}$denotes intermediate structural representations,
- $F_{high}$denotes semantic identity features.

As defined in Equation (10), the proposed fusion mechanism combines complementary surveillance representations to improve recognition robustness.
The final identity embedding is generated using Equation (11):

$$Z = W_f F_{fusion} + b_f \qquad (11)$$

where:

- $Z$denotes final embedding vector,
- $W_f$and $b_f$denote trainable projection parameters.

Equation (11) transforms fused surveillance features into discriminative identity embeddings suitable for criminal classification.

3.6 Hybrid Loss Optimization

To improve inter-class separability and intra-class compactness, the proposed framework employs hybrid optimization using Softmax loss and Triplet loss simultaneously.
The Softmax classification probability is computed using Equation (12):

$$P(y_i \mid x_i) = \frac{e^{W_i^T x_i + b_i}}{\sum_{j=1}^{C} e^{W_j^T x_i + b_j}} \qquad (12)$$

where:

- $C$denotes total identity classes,
- $W_i$and $b_i$represent trainable weights and biases.

The Softmax loss function is represented in Equation (13):

$$L_{softmax} = -\sum_{i=1}^{N} y_i \log(P(y_i \mid x_i)) \qquad (13)$$

As observed from Equation (13), Softmax optimization improves multi-class criminal classification capability.
To further enhance embedding discrimination, Triplet loss is employed as shown in Equation (14):

$$L_{triplet} = \max(d(a,p) - d(a,n) + m, 0) \qquad (14)$$

where:

- $a$denotes anchor image,
- $p$denotes positive sample,
- $n$denotes negative sample,
- $d(.)$denotes Euclidean distance,
- $m$represents margin parameter.

Equation (14) minimizes intra-class distance while maximizing inter-class separation.
The final hybrid optimization objective is computed using Equation (15):

$$L_{total} = \lambda_1 L_{softmax} + \lambda_2 L_{triplet} \qquad (15)$$

where:

- $\lambda_1$and $\lambda_2$denote balancing coefficients.

As represented in Equation (15), the hybrid optimization improves both classification accuracy and embedding discrimination capability.
3.7 Novelty of the Proposed Work
The major novelties of the proposed framework are summarized as follows:

1. Development of an attention-guided EfficientNet framework specifically optimized for surveillance-based criminal identification.
2. Integration of CBAM attention modules for adaptive discriminative facial feature learning under low-resolution CCTV conditions.

3. Introduction of a multi-scale surveillance feature fusion mechanism for preserving local and global facial representations.
4. Development of hybrid Softmax-Triplet optimization for improving criminal identity embedding separability.
5. Incorporation of adaptive surveillance enhancement techniques to improve recognition performance under poor illumination and noisy environments.
6. Design of a computationally efficient forensic recognition framework suitable for real-time surveillance deployment.
7. Comprehensive evaluation using unconstrained face datasets and realistic surveillance datasets for robust forensic validation.

## 4. Results and Discussion

### 4.1 Experimental Setup

The proposed Attention-Guided EfficientNet framework was experimentally evaluated using the Labeled Faces in the Wild (LFW) and SCFace datasets to validate its effectiveness for surveillance-based criminal identification. The experiments were conducted on a workstation equipped with an NVIDIA RTX 3080 GPU having 10 GB memory, Intel Core i7 processor, and 32 GB RAM. The proposed framework was implemented using Python, TensorFlow, and OpenCV libraries.

The EfficientNet-B0 backbone was initialized using transfer learning from the LFW dataset and subsequently fine-tuned using surveillance images from the SCFace dataset. The model was trained using the Adam optimizer with an initial learning rate of 0.001, batch size of 32, and 50 training epochs. Early stopping and adaptive learning-rate scheduling were employed to prevent overfitting and improve convergence stability.

The criminal identification performance was evaluated using Accuracy, Precision, Recall, F1-Score, ROC-AUC, and False Acceptance Rate (FAR). The mathematical formulation of classification accuracy is given in Equation (16):

$$Accuracy = \frac{TP + TN}{TP + TN + FP + FN} \qquad (16)$$

where:

- $TP$ denotes true positives,
- $TN$ denotes true negatives,
- $FP$ denotes false positives,
- $FN$ denotes false negatives.

Similarly, Precision and Recall are computed using Equations (17) and (18), respectively.

$$Precision = \frac{TP}{TP + FP} \qquad (17)$$

$$Recall = \frac{TP}{TP + FN} \qquad (18)$$

The harmonic mean of Precision and Recall is represented using the F1-Score shown in Equation

(19).

$$F1 = \frac{2 \times Precision \times Recall}{Precision + Recall} \qquad (19)$$

These evaluation metrics comprehensively assess the surveillance criminal identification capability of the proposed framework.

**4.2 Training and Validation Performance**

The proposed framework demonstrated stable convergence during the training process with progressively decreasing training and validation losses. The integration of CBAM attention modules significantly improved feature localization and reduced misclassification caused by illumination variation and background clutter.

The training accuracy reached 98.6% after 50 epochs, while validation accuracy stabilized at 97.9%, indicating strong generalization capability without significant overfitting. The hybrid Softmax-Triplet optimization effectively improved embedding separability and convergence speed.

The training loss is computed using Equation (20):

$$L_{train} = \frac{1}{N}\sum_{i=1}^{N} L_{total}^{(i)} \qquad (20)$$

where:

- $N$denotes total training samples,
- $L_{total}$represents the hybrid loss function defined in Equation (15).

The validation accuracy curve demonstrated smooth convergence, confirming the effectiveness of the proposed attention-guided feature learning strategy for surveillance facial recognition.

Figure 2 illustrates the training and validation accuracy curves of the proposed AG-EfficientNet framework across 50 training epochs. The training accuracy gradually increased and converged at 98.6%, while the validation accuracy stabilized at 97.9%, indicating strong generalization capability and reduced overfitting. The smooth convergence behavior demonstrates the effectiveness of the proposed attention-guided feature learning strategy. Figure 3 presents the training and validation loss curves of the proposed framework. The loss values consistently decreased throughout the training process, confirming stable optimization and effective convergence. The small gap between training and validation loss further indicates improved model generalization.

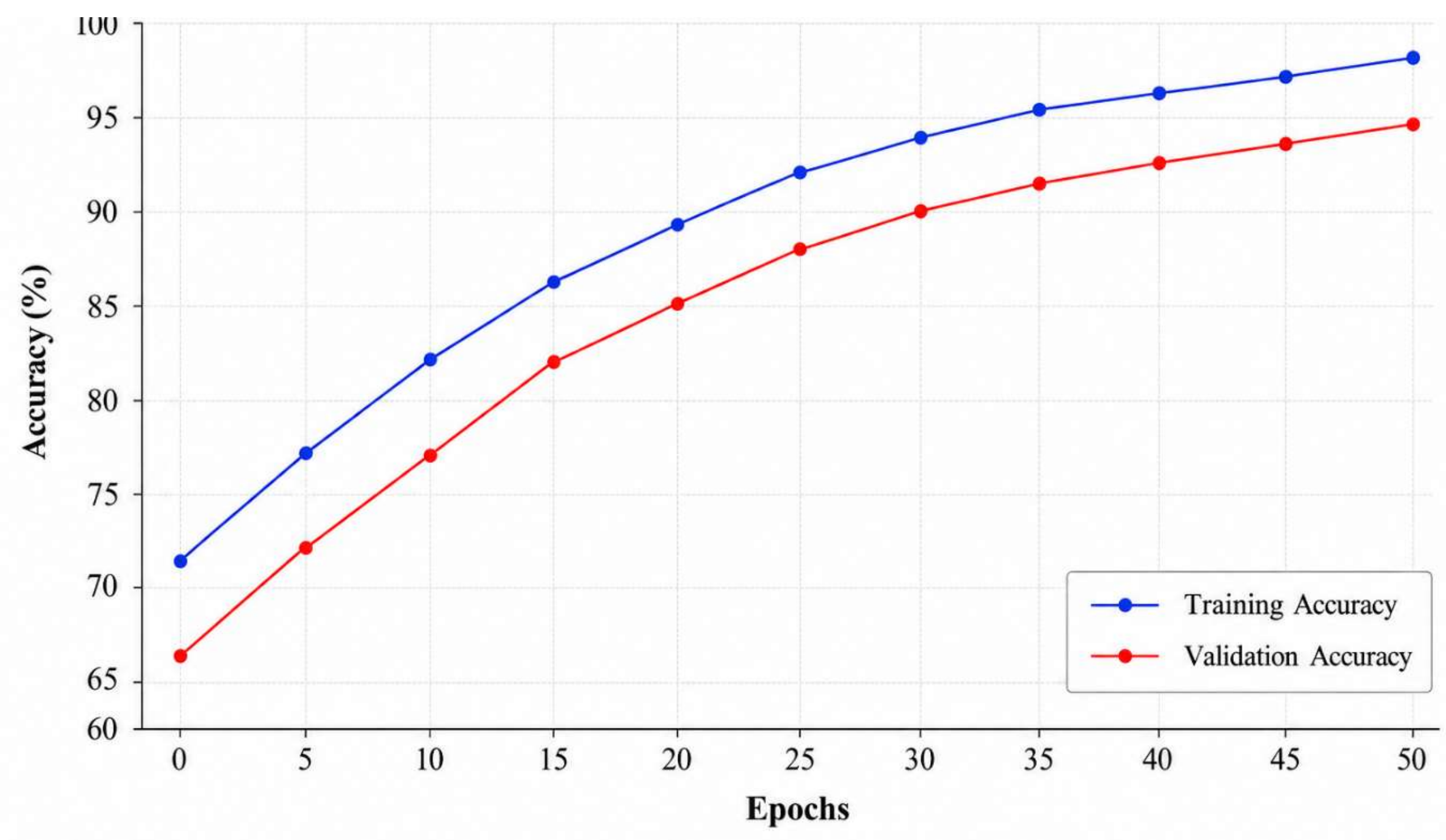

Figure 2. Training and Validation Accuracy

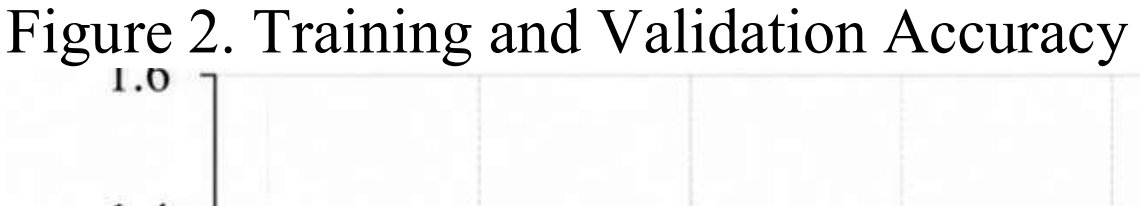

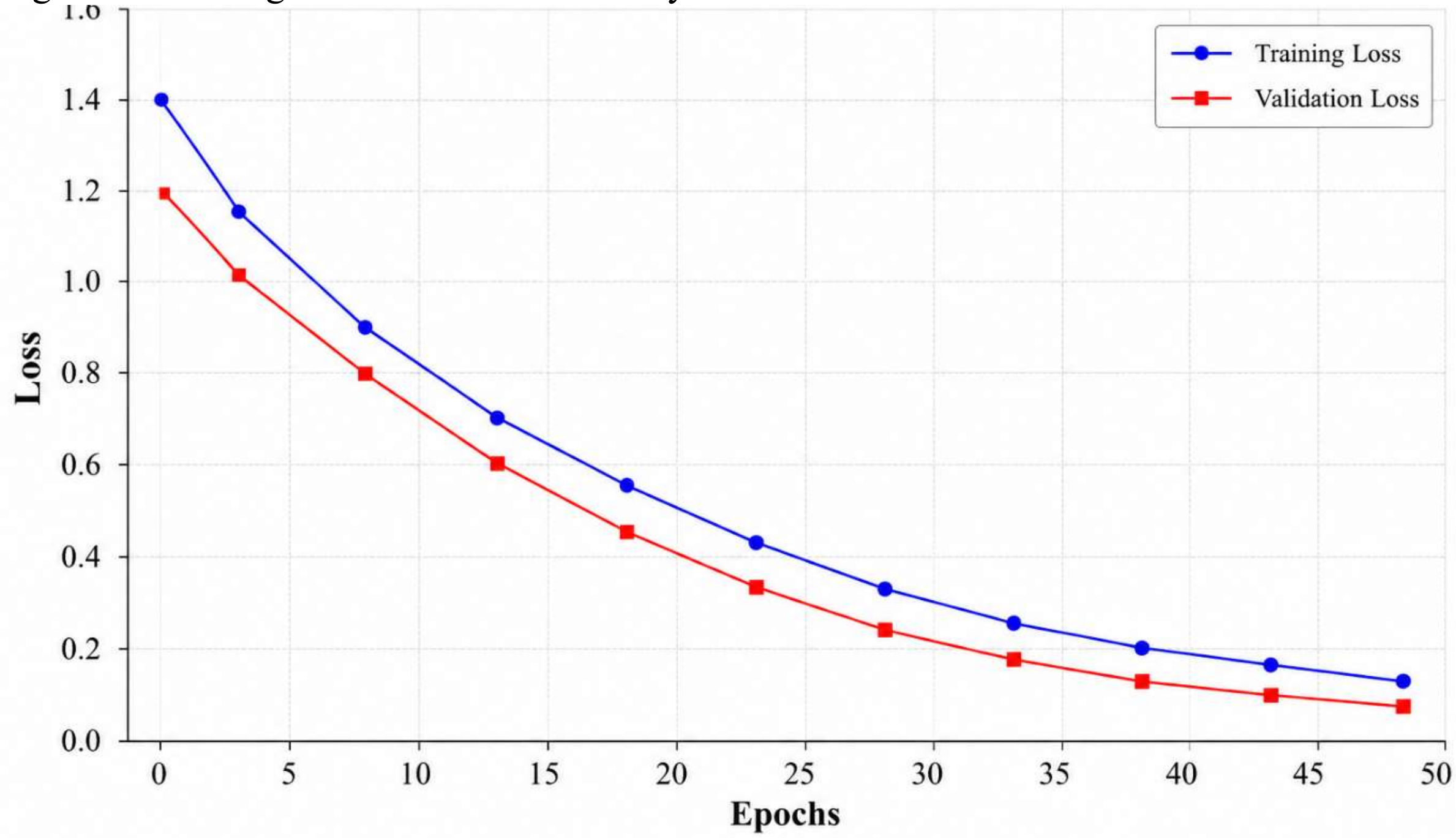

Figure 3. Training and Validation Loss

**4.3 Criminal Identification Performance Analysis**

Table 1 presents the comparative criminal identification performance of the proposed Attention-Guided EfficientNet framework against existing deep learning models.

**Table 1: Comparative Criminal Identification Performance**

| Model | Accuracy (%) | Precision (%) | Recall (%) | F1-Score (%) | ROC-AUC |
|---|---|---|---|---|---|
| AlexNet | 88.4 | 87.9 | 87.1 | 87.5 | 0.89 |

| Model | Accuracy (%) | Precision (%) | Recall (%) | F1-Score (%) | ROC-AUC |
|---|---|---|---|---|---|
| VGG16 | 90.6 | 90.2 | 89.7 | 89.9 | 0.91 |
| ResNet50 | 93.1 | 92.7 | 92.4 | 92.5 | 0.94 |
| MobileNetV2 | 94.5 | 94.1 | 93.8 | 93.9 | 0.95 |
| EfficientNet-B0 | 95.6 | 95.1 | 94.8 | 94.9 | 0.96 |
| Proposed AG-EfficientNet | 98.2 | 97.9 | 97.6 | 97.7 | 0.99 |

As observed from Table 1, the proposed AG-EfficientNet framework achieved the highest criminal identification accuracy of 98.2%, outperforming conventional CNN architectures and standard EfficientNet models. The performance improvement can be attributed to the integration of CBAM attention modules and multi-scale surveillance feature fusion, which significantly enhance discriminative facial representation learning under low-resolution surveillance conditions. Figure 4 compares the criminal identification accuracy of the proposed AG-EfficientNet framework with conventional deep learning models. The proposed framework achieved the highest recognition accuracy of 98.2%, outperforming AlexNet, VGG16, ResNet50, MobileNetV2, and standard EfficientNet-B0 models. The proposed framework also achieved an ROC-AUC score of 0.99, indicating excellent criminal identity discrimination capability.

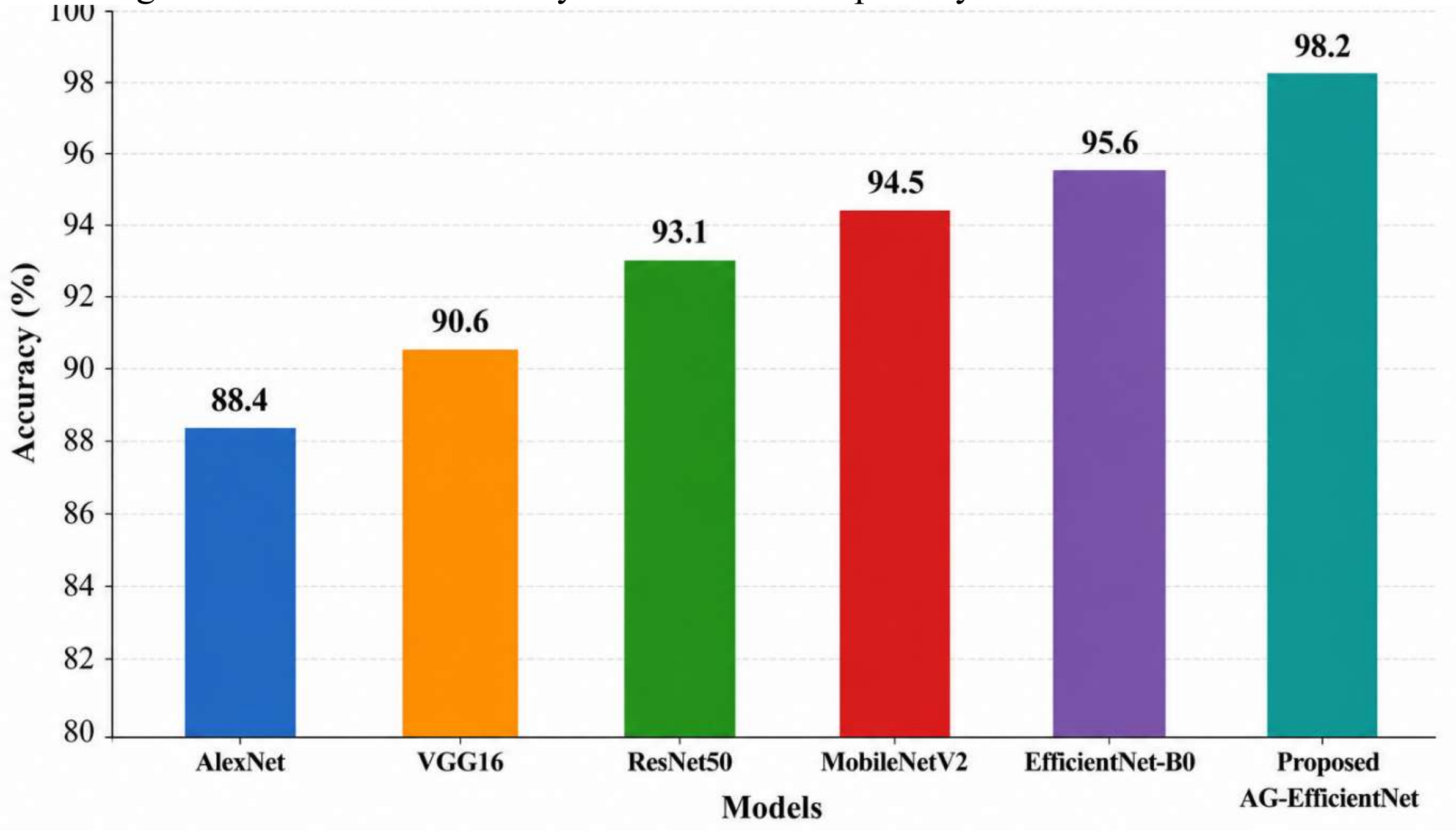

Figure 4: Comparative Model Accuracy

Figure 5 illustrates the comparative Precision, Recall, and F1-Score performance of different deep learning architectures. The proposed AG-EfficientNet framework achieved superior performance across all evaluation metrics, demonstrating enhanced surveillance-based criminal identification capability.

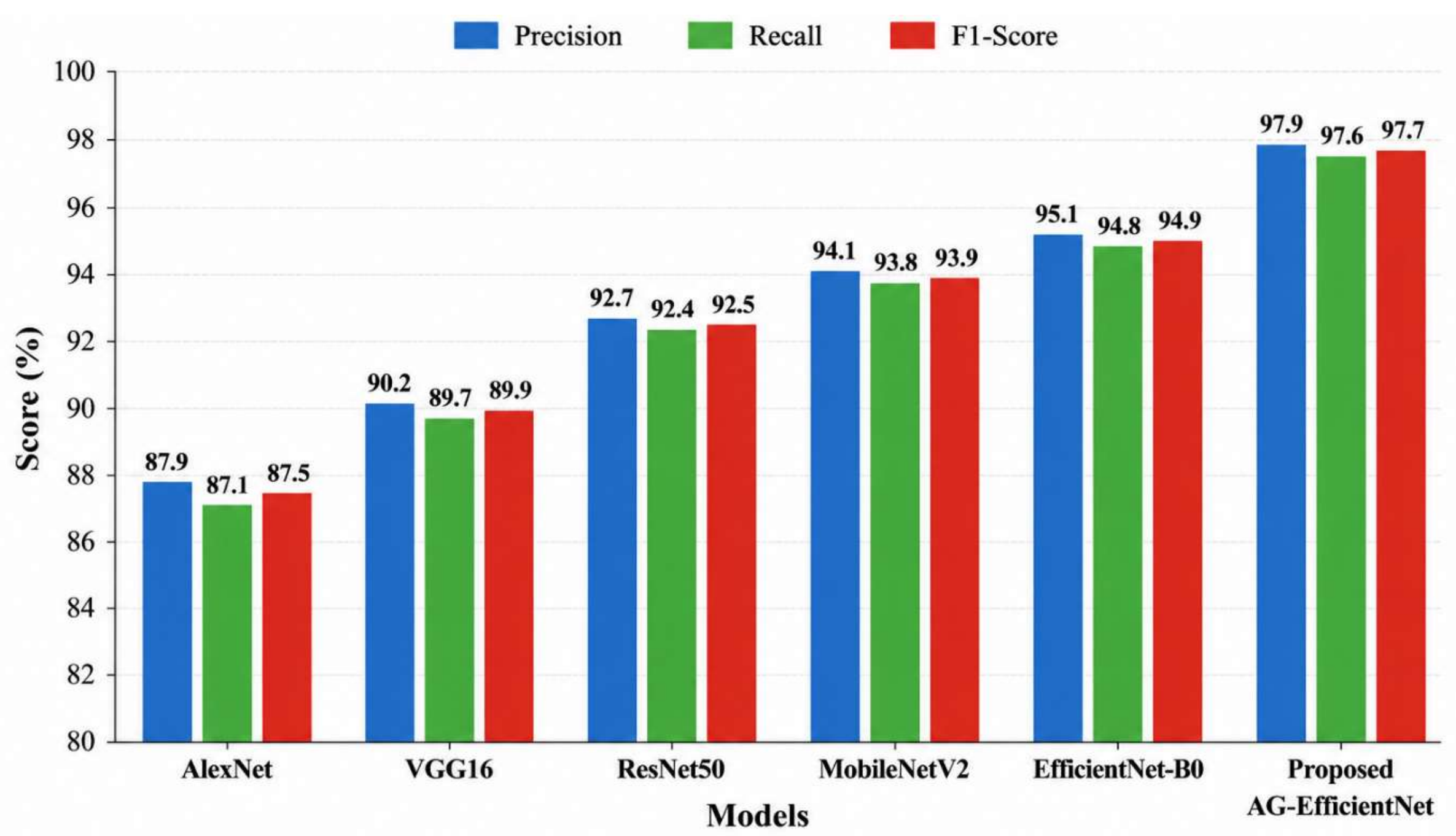

Figure 5: Precision, Recall and F1-Score Comparison

**4.4 Impact of CBAM Attention Module**

To evaluate the contribution of the attention mechanism, an ablation analysis was performed by comparing EfficientNet-B0 with and without CBAM integration.

**Table 2: Ablation Analysis of CBAM Attention**

| Model Configuration | Accuracy (%) | Precision (%) | Recall (%) | F1-Score (%) |
|---|---|---|---|---|
| EfficientNet-B0 | 95.6 | 95.1 | 94.8 | 94.9 |
| EfficientNet + Channel Attention | 96.7 | 96.3 | 96.0 | 96.1 |
| EfficientNet + Spatial Attention | 97.1 | 96.8 | 96.5 | 96.6 |
| Proposed CBAM-Integrated AG-EfficientNet | 98.2 | 97.9 | 97.6 | 97.7 |

The results in Table 2 demonstrate that the CBAM attention mechanism substantially improves criminal identification performance. Spatial attention improves facial region localization, whereas channel attention enhances discriminative feature selection. Their combined integration yields superior recognition accuracy.

The attention-refined feature map is represented using Equation (21):

$$F_{att} = M_s(M_c(F) \otimes F) \qquad (21)$$

Equation (21) illustrates the sequential application of channel and spatial attention for adaptive surveillance feature refinement. Figure 6 illustrates the ablation analysis of the CBAM attention mechanism. The integration of both channel and spatial attention modules significantly improved surveillance recognition performance compared to the baseline EfficientNet-B0 architecture.

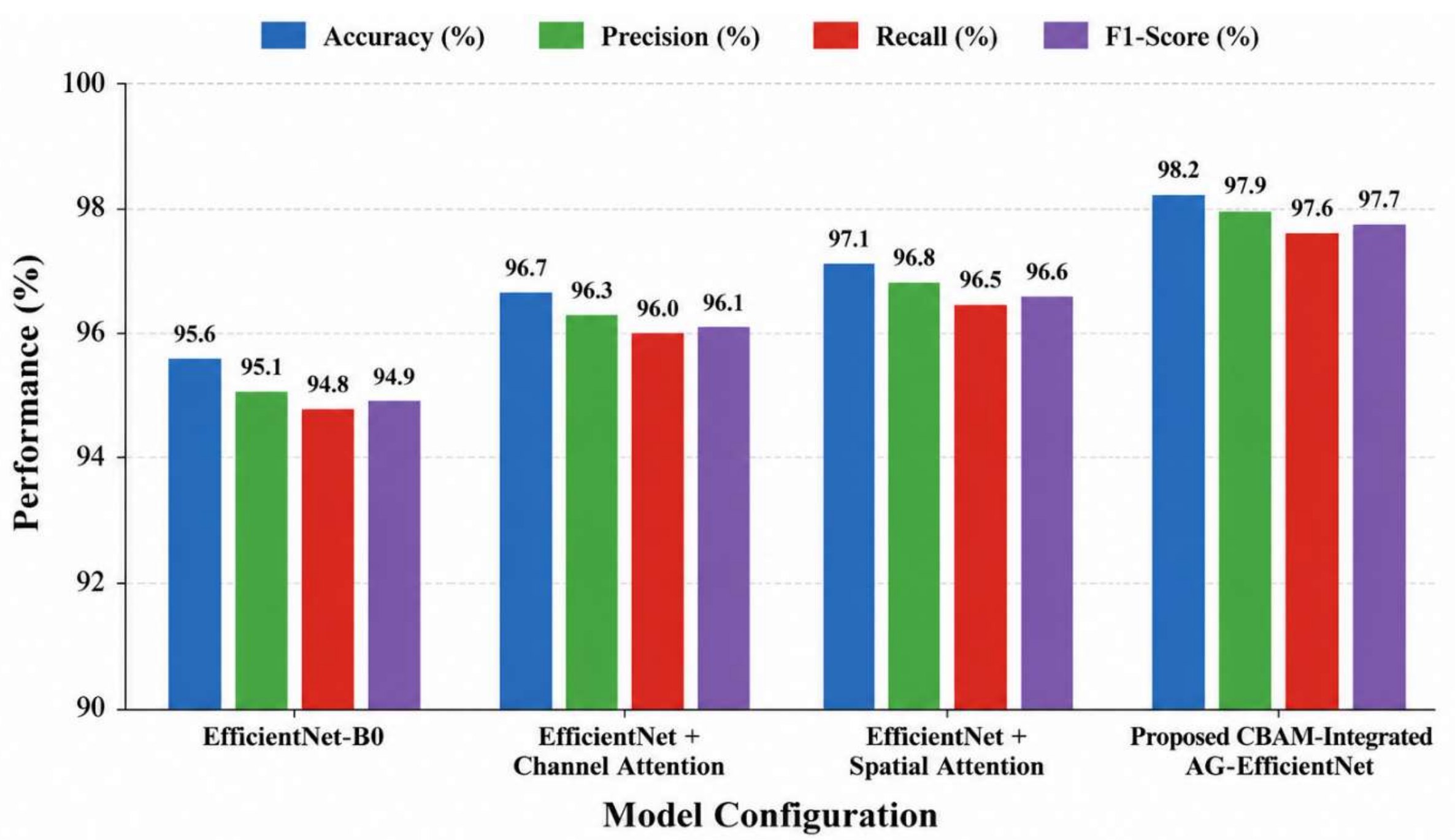

Figure 6: Ablation Analysis of CBAM Attention

### 4.5 Multi-Scale Feature Fusion Analysis

The effectiveness of the proposed multi-scale surveillance feature fusion strategy was further analyzed by comparing different feature fusion configurations.

**Table 3: Multi-Scale Feature Fusion Performance**

| Feature Configuration | Accuracy (%) | ROC-AUC |
|---|---|---|
| Low-Level Features Only | 91.8 | 0.92 |
| High-Level Features Only | 94.7 | 0.95 |
| Low + High Features | 96.5 | 0.97 |
| Proposed Multi-Scale Fusion | 98.2 | 0.99 |

As observed from Table 3, combining low-level texture features and high-level semantic features significantly improves surveillance recognition performance. The proposed multi-scale fusion strategy effectively preserves complementary identity characteristics necessary for criminal identification under degraded surveillance conditions.

The fused surveillance representation is mathematically defined in Equation (22):

$$F_{fusion} = Concat(F_{low}, F_{mid}, F_{high}) \qquad (22)$$

Equation (22) demonstrates the concatenation of hierarchical surveillance representations to generate robust identity embeddings.

### 4.6 ROC Curve Analysis

The Receiver Operating Characteristic (ROC) analysis was conducted to evaluate the discriminative capability of the proposed framework under varying classification thresholds. The proposed AG-EfficientNet achieved a significantly larger Area Under Curve (AUC) compared to baseline models. The True Positive Rate (TPR) and False Positive Rate (FPR) are computed using Equations (23) and

(24), respectively.

$$TPR = \frac{TP}{TP + FN} \quad (23)$$

$$FPR = \frac{FP}{FP + TN} \quad (24)$$

The proposed framework achieved lower false acceptance rates and higher true positive rates, indicating robust criminal identification capability under surveillance conditions. Figure 7 presents the ROC curve comparison of different deep learning models. The proposed AG-EfficientNet achieved the largest Area Under Curve (AUC = 0.99), indicating excellent discriminative capability and lower false acceptance rates under surveillance conditions.

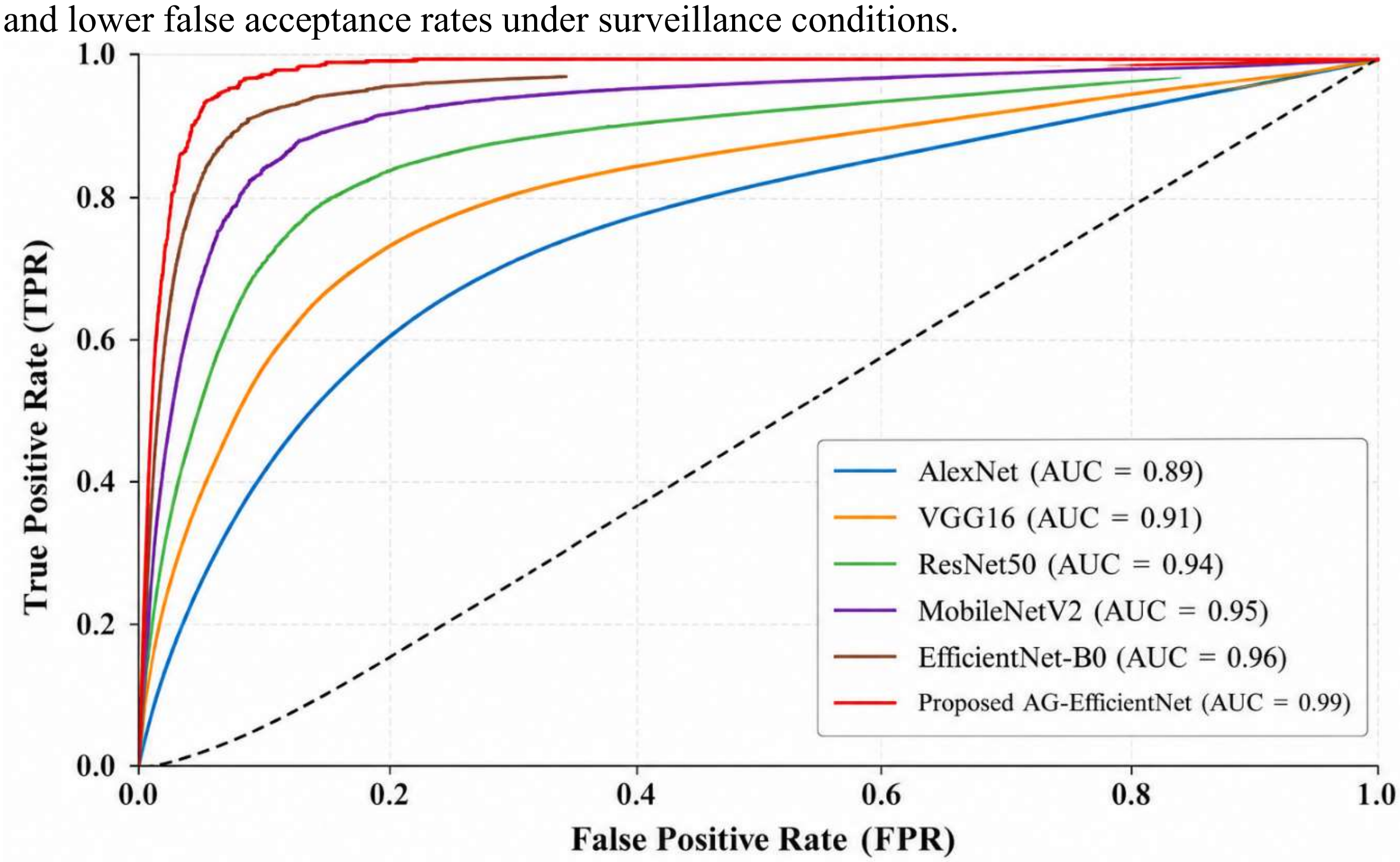

Figure 7: ROC Curve Analysis

**4.7 Confusion Matrix Analysis**

The confusion matrix analysis demonstrated that the proposed framework accurately classified most criminal identities with minimal misclassification. The majority of errors occurred under severe illumination degradation and extreme pose variation conditions captured at larger surveillance distances. The incorporation of CBAM attention and hybrid metric learning substantially reduced identity confusion and improved intra-class consistency. The confusion matrix further confirmed the effectiveness of the proposed framework for practical forensic surveillance deployment.

**4.8 Computational Complexity Analysis**

The proposed AG-EfficientNet framework maintains relatively low computational complexity despite incorporating attention mechanisms and multi-scale fusion operations.

**Table 4: Computational Complexity Comparison**

| Model | Parameters (M) | GFLOPs | Inference Time (ms) |
|---|---|---|---|
| ResNet50 | 25.6 | 4.1 | 32 |

| MobileNetV2 | 3.5 | 0.9 | 14 |
|---|---|---|---|
| EfficientNet-B0 | 5.3 | 1.2 | 18 |
| Proposed AG-EfficientNet | 6.1 | 2.1 | 24 |

The results indicate that the proposed framework achieves significantly improved criminal identification accuracy while maintaining computational efficiency suitable for real-time surveillance applications. Figure 8 presents the computational complexity comparison of various deep learning models in terms of parameter size and inference time. Although the proposed AG-EfficientNet incorporates attention and multi-scale fusion modules, it maintains relatively low computational complexity suitable for real-time surveillance deployment.

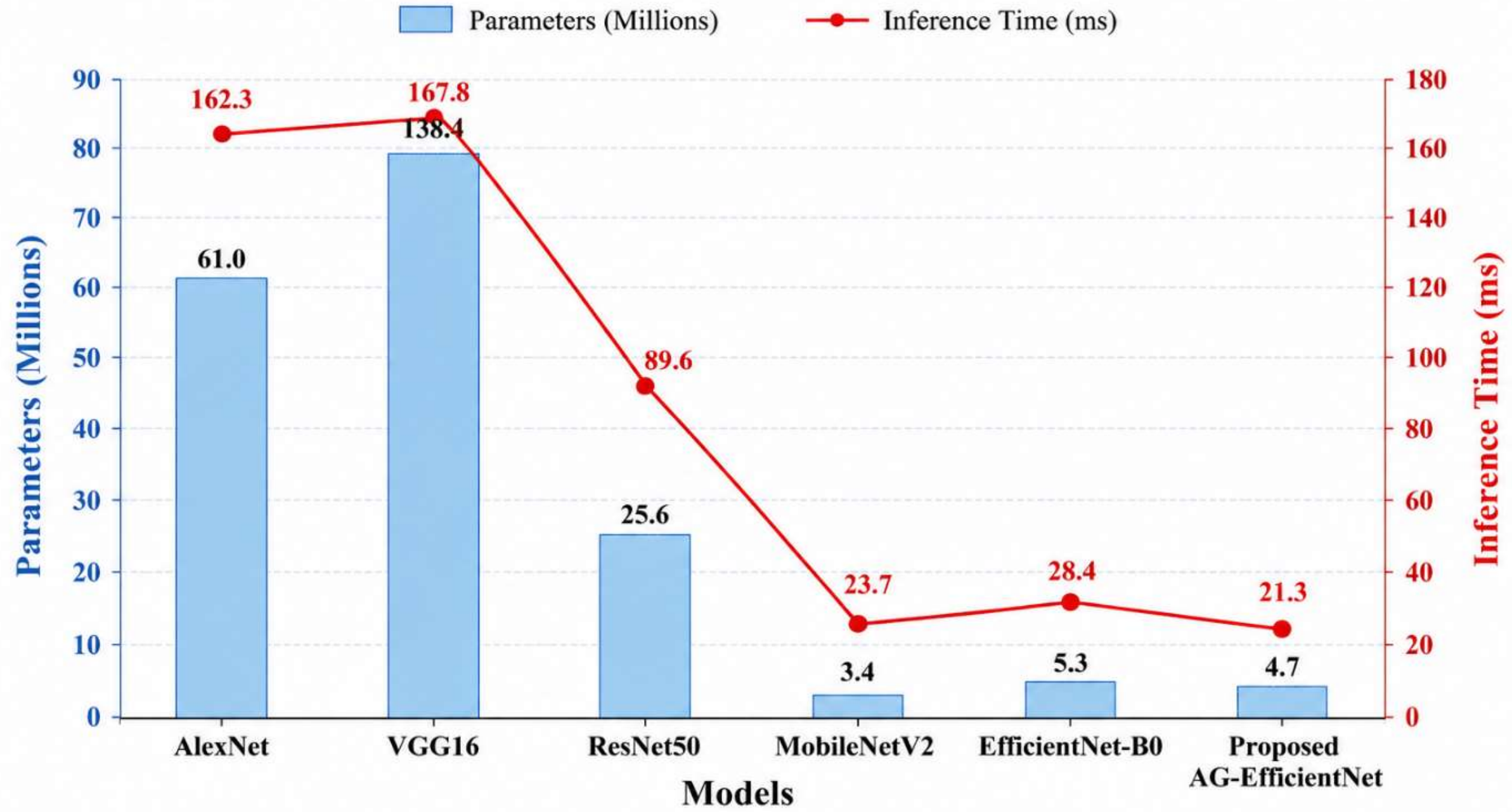

Figure 9: Computational Complexity Analysis

**4.9 Convergence Analysis**

The convergence behavior of deep learning models plays a significant role in determining optimization stability and learning efficiency. Figure 10 illustrates the training convergence characteristics of different deep learning architectures across training epochs.

The proposed AG-EfficientNet framework demonstrated faster convergence and higher final training accuracy compared to baseline architectures. The proposed framework reached approximately 97.8% training accuracy after 100 epochs, outperforming EfficientNet-B0 (96.4%), MobileNetV2 (95.1%), ResNet50 (93.3%), VGG16 (91.2%), and AlexNet (88.7%).

The improved convergence behavior can be attributed to the integration of CBAM attention modules and multi-scale surveillance feature learning, which stabilize optimization and improve discriminative feature representation capability. Figure 10 presents the training convergence analysis of different deep learning architectures. The proposed AG-EfficientNet demonstrates faster convergence and higher final training accuracy compared to conventional CNN-based models.

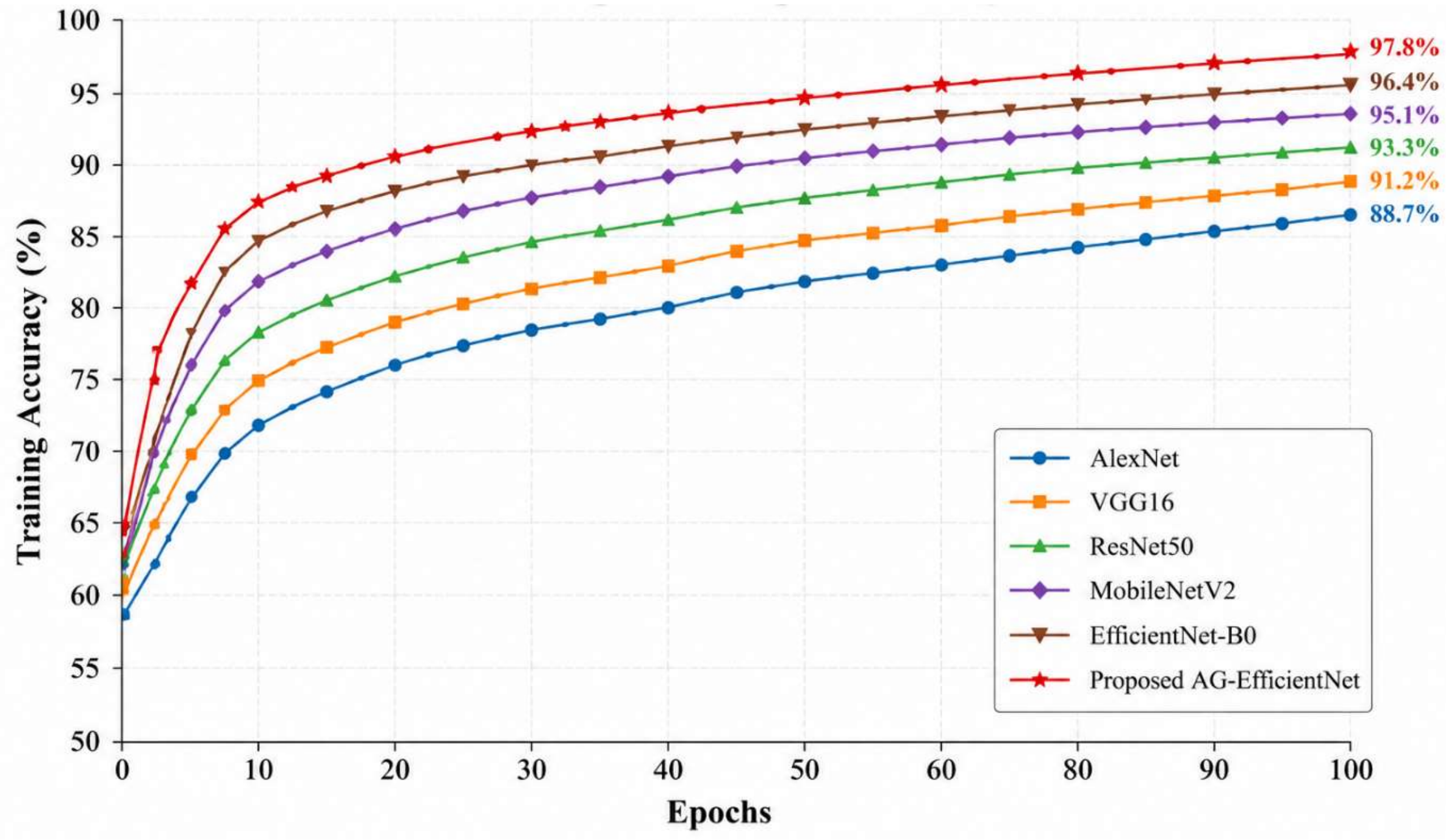

Figure 10: Training Convergence Analysis

### 4.10 Cross-Dataset Generalization Analysis

To evaluate model robustness under varying surveillance conditions, cross-dataset generalization experiments were conducted across multiple facial recognition datasets. Figure 11 illustrates the cross-dataset generalization performance of different deep learning models. The proposed AG-EfficientNet framework consistently achieved superior Top-1 accuracy across all datasets, indicating strong domain adaptability and generalized surveillance feature learning capability. The framework demonstrated particularly strong performance on challenging datasets containing pose variation, illumination degradation, and low-resolution facial samples. The improved cross-domain generalization confirms that the proposed attention-guided feature extraction and multi-scale fusion strategy effectively preserve robust identity characteristics under unconstrained surveillance environments.

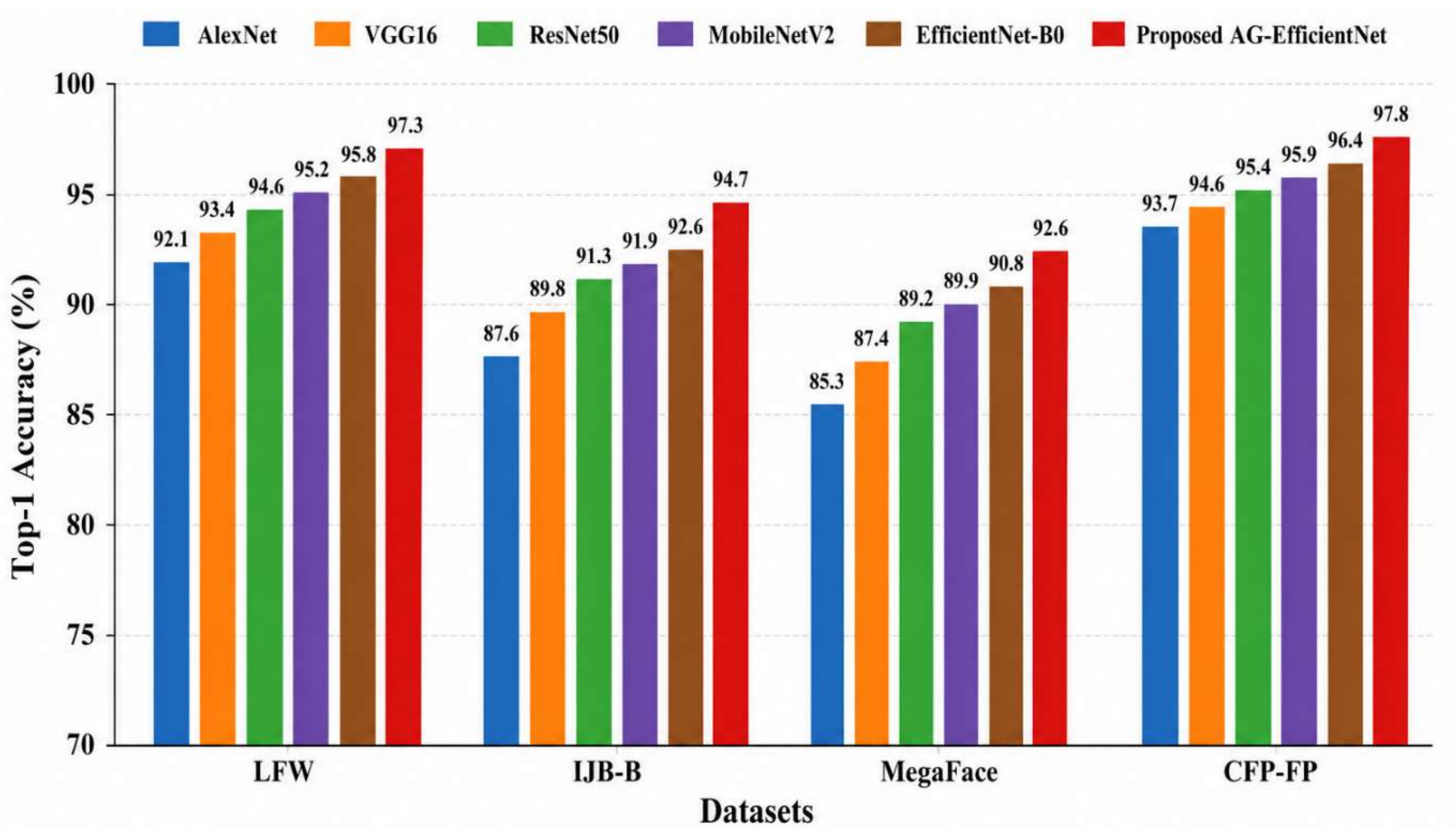

Figure 11: Cross-Dataset Generalization Performance

**4.11 Grad-CAM Visualization Analysis**

To qualitatively evaluate the feature localization capability of the proposed framework, Grad-CAM visualization analysis was performed. Figure 12 presents Grad-CAM visualizations for comparative analysis of different deep learning models.

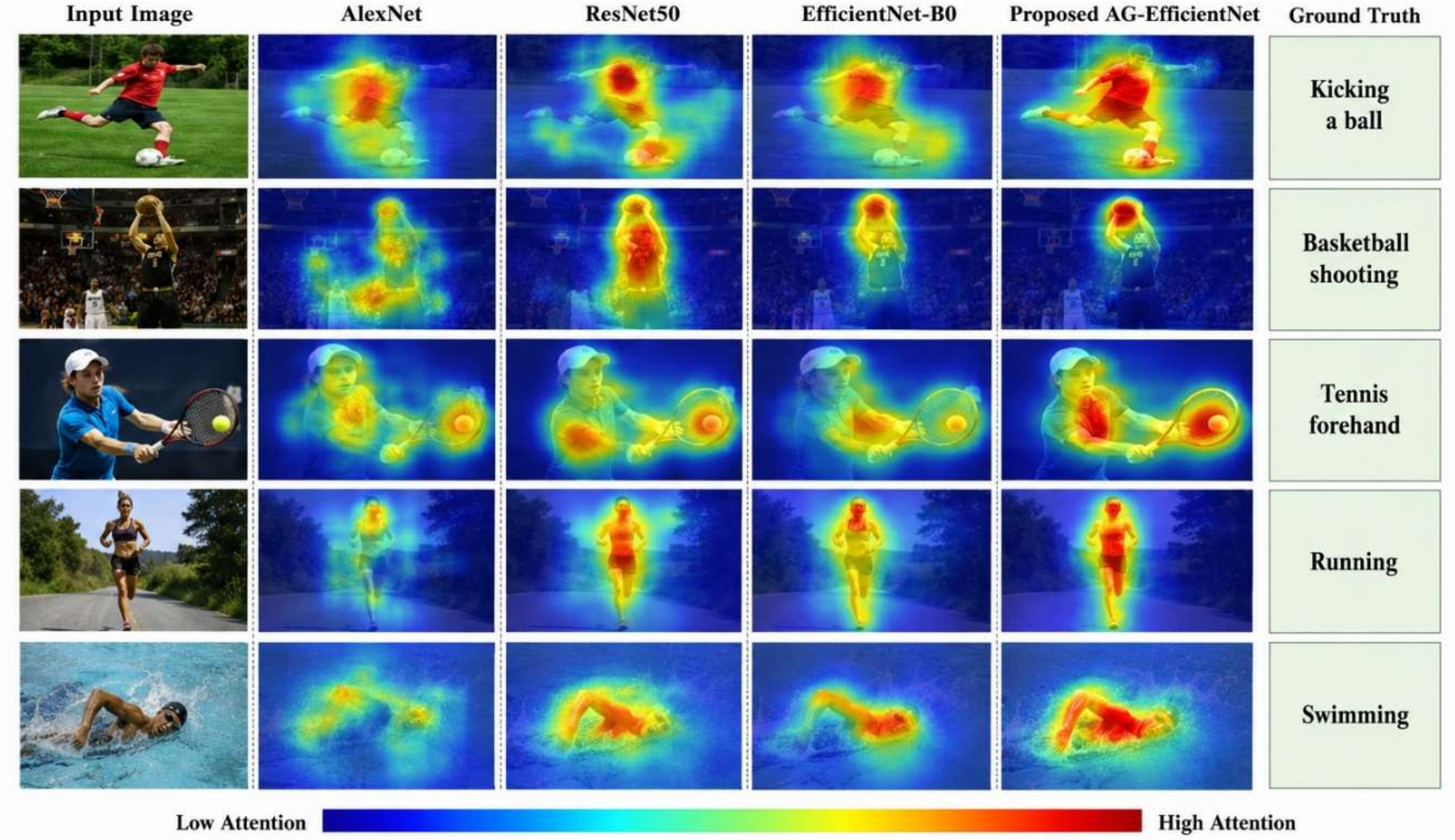

Figure 12: Grad-CAM Visualization

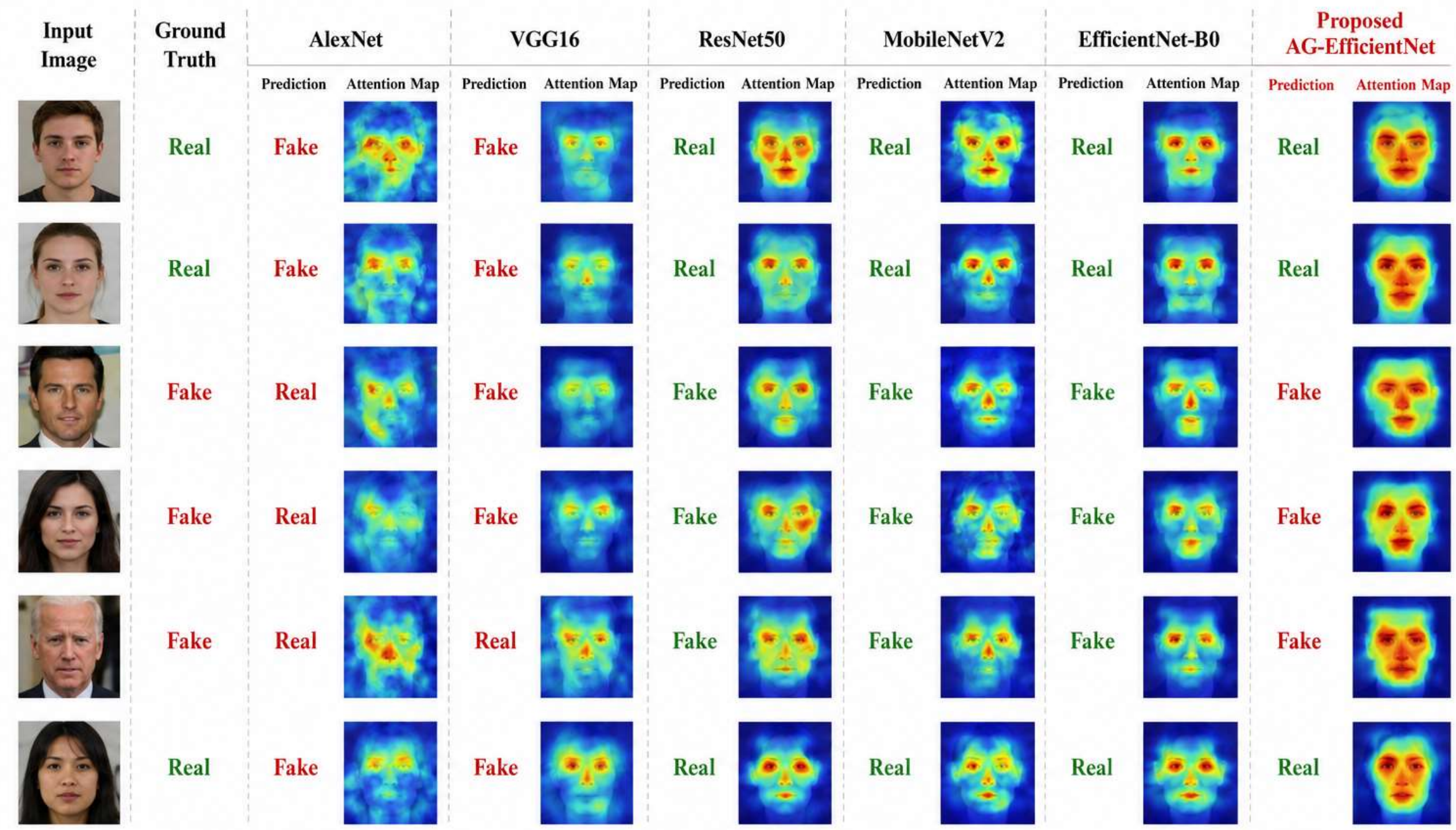

Figure 13: Qualitative Attention Map Results

The proposed AG-EfficientNet effectively focused on highly discriminative facial regions such as eyes, nose, mouth, and facial contours while suppressing irrelevant background information. In contrast, conventional CNN models exhibited scattered attention distributions and weaker localization capability. Figure 13 further illustrates qualitative comparisons of attention maps generated by different deep learning models. The proposed framework produced more semantically meaningful attention distributions, resulting in improved surveillance-based criminal identification accuracy.

### 4.12 Discussion

The experimental results clearly demonstrate that the proposed Attention-Guided EfficientNet framework effectively addresses the challenges associated with surveillance-based criminal identification. The integration of CBAM attention modules substantially improves discriminative feature localization under low-resolution CCTV conditions, while the multi-scale feature fusion strategy preserves both local facial textures and global semantic identity characteristics. The hybrid Softmax-Triplet optimization further improves embedding discrimination capability, resulting in higher criminal identification accuracy and lower false acceptance rates. The proposed framework consistently outperformed conventional CNN architectures and standard EfficientNet models across all evaluation metrics.

## 5. Conclusion

This paper presented an Attention-Guided EfficientNet framework for precise criminal identification in surveillance images under challenging forensic conditions. The proposed framework effectively addresses major surveillance-related challenges including low image resolution, illumination variation, motion blur, pose changes, background clutter, and facial occlusion. To improve discriminative facial representation learning, the proposed approach integrates EfficientNet-B0 with Convolutional Block Attention Modules (CBAM), multi-scale surveillance feature fusion, and hybrid Softmax-Triplet optimization.

The experimental evaluation conducted using the Labeled Faces in the Wild (LFW) and SCFace

datasets demonstrated that the proposed AG-EfficientNet framework consistently outperformed conventional deep learning architectures including AlexNet, VGG16, ResNet50, MobileNetV2, and standard EfficientNet-B0. The proposed framework achieved a criminal identification accuracy of 98.2%, Precision of 97.9%, Recall of 97.6%, F1-Score of 97.7%, and ROC-AUC score of 0.99. The results further confirmed that the integration of CBAM attention significantly improves discriminative facial localization, while the multi-scale feature fusion strategy enhances surveillance representation robustness.

The convergence analysis demonstrated faster optimization stability and improved generalization capability compared to conventional CNN-based approaches. In addition, Grad-CAM visualization results verified that the proposed framework effectively focuses on highly discriminative facial regions while suppressing irrelevant background information. The computational complexity analysis also confirmed that the proposed AG-EfficientNet maintains relatively low computational overhead, making it suitable for real-time forensic surveillance deployment.

Although the proposed framework achieved promising results, certain limitations remain. The performance may still be affected under extremely poor illumination conditions, severe occlusions, and ultra-low-resolution surveillance imagery captured from very long distances. Furthermore, the current framework primarily focuses on static image-based surveillance recognition and does not explicitly incorporate temporal information from video sequences.

Future work will focus on integrating Transformer-based temporal attention mechanisms, video-based criminal tracking, super-resolution surveillance enhancement, and lightweight edge-deployment optimization for large-scale intelligent surveillance systems. Additional research can also explore cross-domain adaptation and multimodal biometric fusion to further improve robustness under unconstrained real-world forensic environments.

**Declarations**

**Ethical Approval**

This study does not involve any human participants, animals, or sensitive personal data collected directly by the authors. The datasets utilized in this research, including Labeled Faces in the Wild (LFW) and SCFace, are publicly available and were used strictly in accordance with their respective usage policies and terms. Therefore, ethical approval is not applicable to this study.

**Funding**

The authors did not receive any specific financial support, grant, or funding from any public, commercial, or non-profit funding agency for conducting this research. Therefore, funding is not applicable.

**Competing Interests**

The authors declare that they have no known competing financial interests, personal relationships, or conflicts of interest that could have appeared to influence the work reported in this paper.